\documentclass[letterpaper, 10 pt, conference]{ieeeconf}  

\IEEEoverridecommandlockouts                              

\usepackage{dblfloatfix} 

\usepackage{listings}

\usepackage{capt-of}    
\usepackage{etoolbox}   

\usepackage{graphicx}
\usepackage{booktabs}
\usepackage{multirow}
\usepackage{array}
\usepackage{cuted}
\usepackage{amsmath} 
\usepackage{cuted}
\usepackage{afterpage}
\usepackage{capt-of} 
\usepackage{etoolbox} 

\usepackage{amsmath}
\usepackage{amsfonts}
\usepackage{makecell}
\usepackage{tabularx}
\usepackage{vcell}
\usepackage{wrapfig,lipsum,booktabs}
\usepackage{multirow}
\usepackage{xspace}
\usepackage{caption}
\usepackage{subcaption}
\usepackage{wrapfig}

\usepackage{booktabs}      
\usepackage{graphicx}      
\usepackage{hyperref}  
\usepackage{xspace}
\usepackage{siunitx}

\usepackage[table,xcdraw,dvipsnames]{xcolor}

\usepackage[para]{threeparttable}
\usepackage{tabularx}
\usepackage{booktabs}
\usepackage{dsfont}
\usepackage{bm}
\usepackage{float}

\usepackage{array}
\usepackage{booktabs}
\usepackage{xcolor,colortbl}
\usepackage{ragged2e}

\title{\LARGE \bf
FAME: Force-Adaptive RL for Expanding the Manipulation Envelope of a Full-Scale Humanoid
}

\author{
Niraj Pudasaini$^{1}$,
Yutong Zhang$^{2}$,
Jensen Lavering$^{2}$,
Max Conway$^{2}$,
Alessandro Roncone$^{2}$,
and Nikolaus Correll$^{1}$%
\thanks{$^{1}$Department of Aerospace and Mechanical Engineering, University of Notre Dame, Notre Dame, IN 46556, USA.}%
\thanks{$^{2}$Department of Computer Science, University of Colorado Boulder, Boulder, CO 80309, USA.}%
\thanks{Corresponding author: Niraj Pudasaini, \texttt{npudasai@nd.edu}.}
}
\begin{document}

\maketitle
\begin{strip}
\vspace{-2.7cm}
\centering
\includegraphics[width=\textwidth]{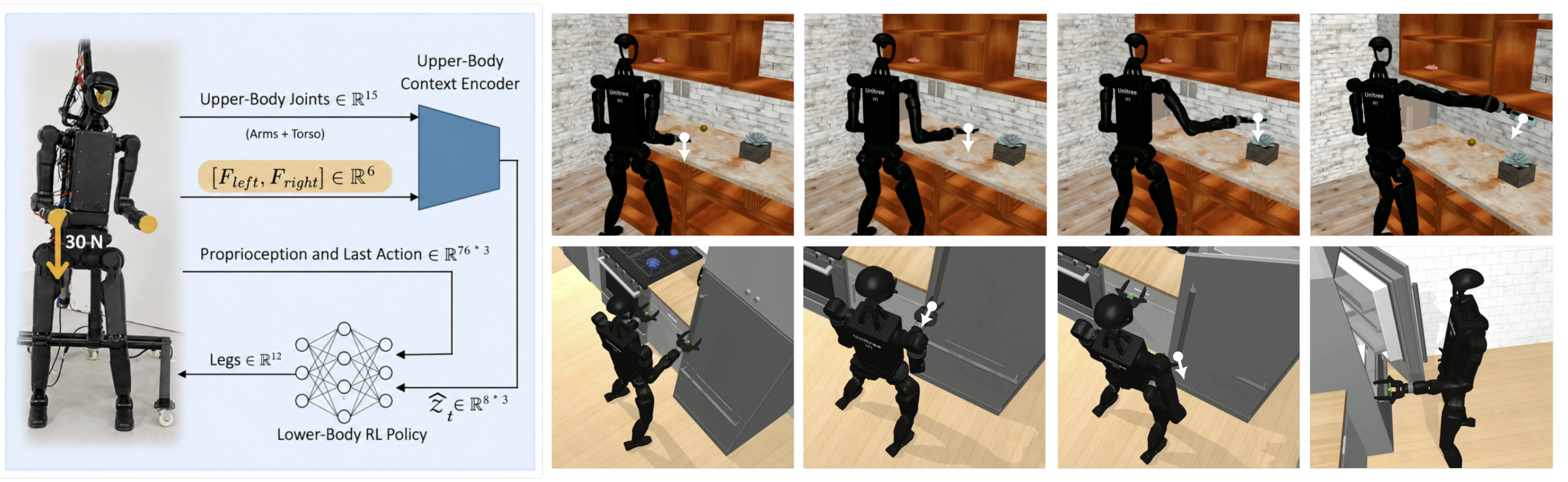}
\vspace{-5mm}
\captionof{figure}{FAME overview. \textbf{Left:} a standing policy conditioned
on an upper-body context encoder mapping arm and torso configuration $\in
\mathbb{R}^{15}$ and hand forces $[F^L, F^R] \in \mathbb{R}^{6}$ to a latent
context $z_t$. At deployment the forces are reconstructed from joint torques. The H1-2 stands under a
$30$\,N load. \textbf{Right:} the policy in the GOLEM
simulation~\cite{conway2026golemmodularhumanoidautonomy} on bowl pick-and-place (top)
and refrigerator door opening (bottom), where the forces arise from the task.
Arrows mark the force on the hand.}
\label{fig:overview}
\vspace{-4mm}
\end{strip}

\thispagestyle{empty}
\pagestyle{plain}

\begin{abstract}
Maintaining balance under external hand forces is critical for humanoid bimanual
manipulation, where interaction forces propagate through the kinematic chain and
constrain the feasible manipulation envelope. We propose FAME, a force-adaptive
reinforcement learning framework that conditions a standing policy on a learned
latent context encoding upper-body joint configuration and bimanual interaction
forces \emph{jointly}, since the base moment a load induces depends on the arm
configuration through which it acts. Training applies isotropically sampled 3D
forces at each hand under an upper-body pose curriculum, exposing the policy to
manipulation-induced perturbations across continuously varying arm
configurations. At deployment the interaction force is not measured but
reconstructed online from joint torques and states through rigid-body inverse
dynamics, requiring no wrist force/torque sensing. We evaluate over $100$
upper-body configurations under swept hand forces, scoring each trial by a
task-level criterion that requires the robot both to remain upright and to hold
its hands near where the task placed them; all such results run with the
estimated force in the loop. At a $150$\,mm tolerance FAME reaches $38.9\%$ task
success, against $16.6\%$ for a policy given the same force without encoding,
$4.3\%$ for a pose-conditioned curriculum policy, and $24.7\%$ for an
adversarially trained locomotion policy, which stays upright but recovers by
stepping and so relocates the hands. We further demonstrate transfer to
task-generated interaction forces in a MuJoCo kitchen environment, and to
asymmetric and bimanual loading on a full-scale Unitree H1-2. Code and videos are available on the \href{https://correlllab.github.io/fame_website/}{project website}.
\end{abstract}



\section{INTRODUCTION}
\label{sec:introduction}
Humanoid robots are designed to operate in human-centered environments, where interaction with tools, objects, and existing infrastructure requires coordinated upper-body bimanual manipulation with stable lower-body control. During bimanual manipulation, external forces applied at the hands propagate through the kinematic chain and directly disturb the lower-body balance. Unless actively mitigated, these interaction forces constrain the feasible manipulation envelope of the robot and couple upper-body tasks with stance stability. We define the manipulation envelope as the region of admissible external hand forces and arm configurations under which the robot can maintain stable standing. Enabling humanoids to maintain stable standing under varying and uncertain hand loads remains a fundamental challenge.

\begin{figure*}[!t]
  \centering
  \includegraphics[width=17cm]{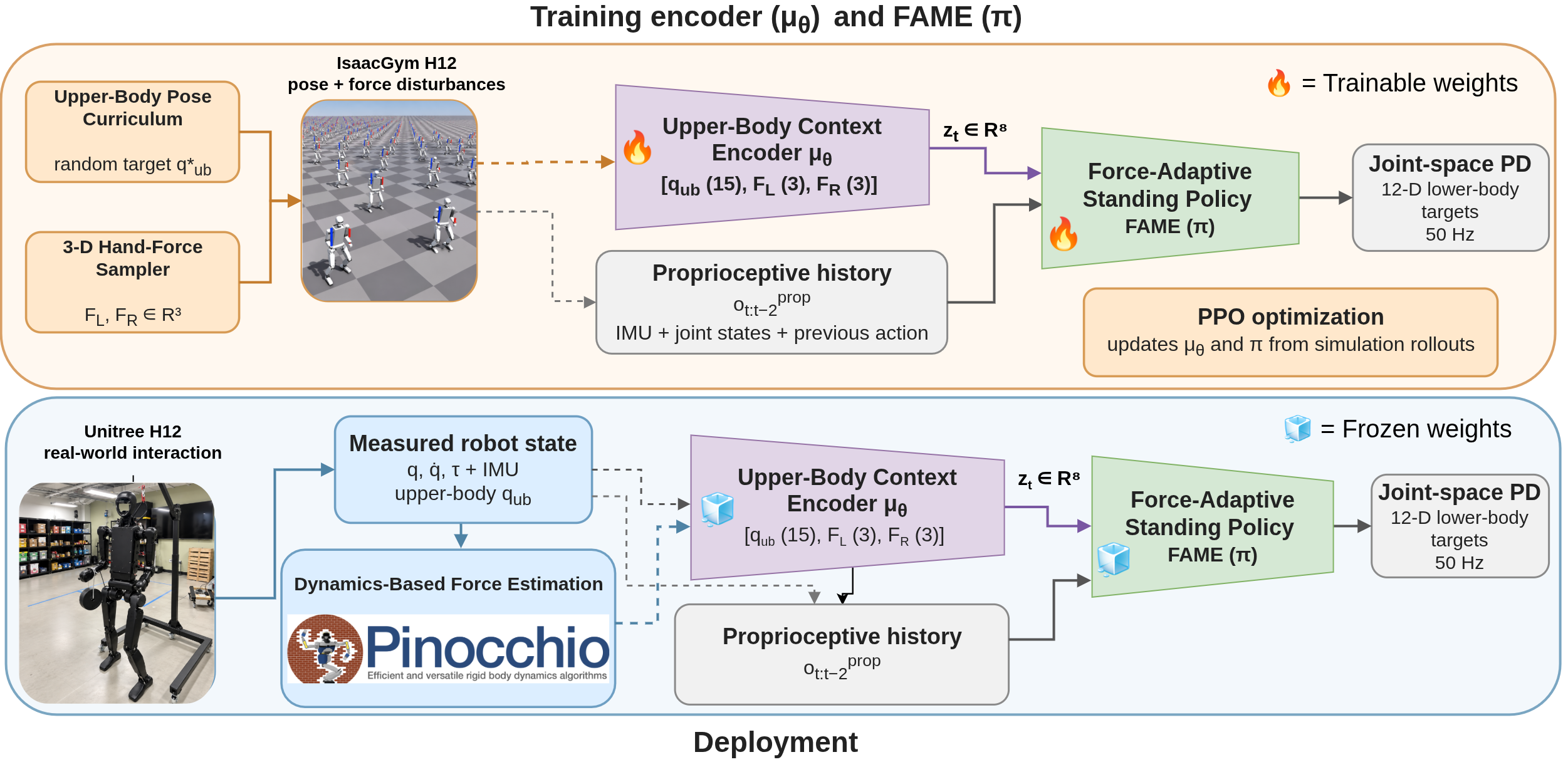}
 \caption{ Overview of the proposed standing framework. 
During training (top), an upper-body dynamics encoder processes upper-body joint states and sampled hand forces to produce a latent context variable that conditions the base standing policy in simulation. 
During deployment (bottom), the same encoder operates on measured upper-body joints and estimated hand forces to infer the latent context online, enabling rapid adaptation to upper-body-induced disturbances.\label{fig:methodology}}
\end{figure*}

Traditional model-based methods~\cite{Gupta18062017,reher2020dynamicwalkingagileefficient}
(LIPM~\cite{kajita2001lipm}, SLIP~\cite{6697099}, MPC, trajectory optimization) for humanoid balance struggle under highly dynamic conditions with significant external disturbances. Deep reinforcement learning (DRL) has emerged as a promising alternative \cite{schulman2017proximalpolicyoptimizationalgorithms,gu2024humanoid,heess2017emergencelocomotionbehavioursrich}, yet training policies that maintain stability under varying and uncertain hand forces during manipulation remains challenging. The core difficulty lies in the high-dimensional, coupled nature of the disturbance: interaction forces at the hands depend on both the magnitude and direction of applied loads as well as the geometric configuration of the arms, creating a state space that is difficult to explore during training.

Latent context adaptation supplies the right machinery, though not for the usual
reason. Rapid Motor Adaptation (RMA)
\cite{kumar2021rma,9981091}
compresses environment parameters such as terrain, mass distribution and contact
dynamics into a latent, and because those parameters cannot be measured on
hardware, regresses that latent from proprioceptive history in a second stage. Domain randomization \cite{8202133} forgoes
the latent and buys robustness by exposure alone. Hand forces are not of that
kind. The robot neither commands the load nor can predict it from its own action
history, yet the load is recoverable at run time from the rigid-body model, so
the latent need not stand in for an unmeasurable quantity and no second-stage
regression is required. Recent humanoid work either exposes the policy to
disturbance without representing it, through upper-body pose curricula
\cite{BenQ-RSS-25} and force-aware training
\cite{zhang2025falcon}, or obtains it explicitly, through co-trained estimators
of a base-applied force fed directly to the policy
\cite{10.1007/978-981-95-2095-4_5} and wrist-mounted force/torque sensing
\cite{11203041}. Each route
either requires dedicated force sensing at deployment or conditions on a
disturbance whose effect on the base is fixed by a known application point.
Forces applied at the hands are different: the induced base moment depends
jointly on the force and on the arm configuration through which it acts, so
neither quantity alone determines the disturbance and the two must be
represented together.

The contributions of this paper are:
\begin{enumerate}
    \item We propose \textbf{FAME}, a force-adaptive standing framework that conditions a lower-body policy on a latent context encoding upper-body
    configuration and bimanual interaction force \emph{jointly}, rather than
    supplying the policy with two independent signals.
    
    \item We \textbf{deploy without force sensing}. The interaction force is not
    privileged but reconstructed online from measured joint torques and states
    through rigid-body inverse dynamics, requiring no wrist force/torque sensing,
    no force supervision, and no base-wrench measurement. The estimator itself is
    standard; what we show is that the policy's advantage survives substituting the
    estimate for the applied force.

    \item  We evaluate over $100$ upper-body configurations sampled from the training pose distribution under swept hand forces, scoring each trial by a \textbf{task-level criterion} that requires the hands to stay near their commanded pose. Against a policy given the same estimated force without the joint encoding, task success rises from $16.6\%$ to $38.9\%$, attributing the gain to the encoding rather than to the availability of the force signal. We further benchmark against \textbf{HOMIE}-style pose-curriculum training~\cite{BenQ-RSS-25} and
    \textbf{ALMI}~\cite{shi2025almi}, an adversarially trained whole-body controller.
       
    \item We validate FAME on the \textbf{Unitree H1-2} full-scale humanoid under representative load-interaction scenarios, including asymmetric single-arm  pulling (RE1) and symmetric bimanual load augmentation (RE2), and evaluate on manipulation tasks in kitchen environments where the interaction forces arise from the task itself.
\end{enumerate}


\section{Related Work}
\label{sec:related_work}

\subsection{Force-adaptive humanoid control}
Recent whole-body control research addresses stability under external
interaction forces. FALCON \cite{zhang2025falcon} trains a dual-agent
policy with a torque-limit-aware 3D force curriculum for forceful tasks
such as heavy lifting and cart-pulling. HAFO
\cite{DONG2026106978} models disturbances
using virtual spring-damper systems, Thor \cite{Li2025ThorTH} introduces
force-adaptive torso-tilt rewards to maximize peak interaction strength,
and Heterogeneous Meta-Control \cite{wei2025hmclearningheterogeneousmetacontrol}
stitches position, impedance, and hybrid force-position modalities within
a shared torque-space interface. These frameworks optimize how much force
the robot can \emph{apply} during loco-manipulation. FAME targets the
complementary quantity: the region of external hand loads and arm
configurations under which stance remains stable, which bounds what
manipulation can be attempted at all.

\subsection{Obtaining interaction forces at deployment}
Conditioning a policy on interaction force requires that force to be available
at run time, and prior work takes two routes. Sensor-based methods instrument
the contact: H2-COMPACT
\cite{11203041} reads
wrist-mounted force/torque sensors to infer human intent during human--humanoid
co-manipulation and adapt contact trajectories online, at the cost of dedicated
force sensing on the robot. Learning-based methods infer force from
proprioception: RL-Force \cite{10.1007/978-981-95-2095-4_5} co-trains an MLP
estimator with the policy to regress a base-applied force vector and its point
of application from a short proprioceptive history, feeding the estimate
directly into the policy. The authors report two failure modes of this design.
Transplanting the estimator into a separately trained policy destabilized both,
indicating that it had learned a mapping valid only for the specific
closed-loop dynamics it was trained inside. Further, the estimated force tracked
the robot's torso pitch closely enough to be largely predictable from it,
suggesting that the estimator had learned to read the postural deflection the
force produces rather than the force itself, which holds only while the policy
responds to disturbances like training.

FAME takes a third route, recovering the external force at each hand from the
difference between modelled and measured joint torques through the wrist
Jacobian, with the model evaluated by full inverse dynamics rather than a static
approximation. This requires no wrist sensing, no force supervision, and no
co-training, and because the mapping follows from the robot model rather than
from data collected under a particular controller, the estimator is independent
of the policy it serves. It is also independent of contact: formed from the rows
of the equation of motion belonging to a single arm, it drops the unknown
ground reaction identically rather than modelling or measuring it, and it needs
no base-wrench measurement, which a physical robot does not have. Both
properties are derived in Sec.~\ref{subsec:force_estimation}.

\subsection{Conditioning on disturbance context}
Rapid Motor Adaptation \cite{kumar2021rma} learns a latent
encoder of environment parameters in simulation and, because those parameters
cannot be measured on hardware, trains a second-stage module to regress the
latent from proprioceptive history; the paradigm has been extended to bipedal
locomotion \cite{9981091} and to manipulators
\cite{10655557}. A separate line forgoes explicit
conditioning and relies on randomized exposure, with humanoid controllers
training under continuously varying arm configurations via upper-body pose
curricula \cite{BenQ-RSS-25}, or
adversarially, as in ALMI \cite{shi2025almi}, which trains a locomotion policy
against upper-body motions ordered by how much they destabilize the current
policy. Both follow broader use of curricula for command ranges, terrain
difficulty, and perturbation magnitude in legged locomotion
\cite{Hwangbo_2019,rudin2021learning}.

The two lines differ from our setting in what the policy can know. RMA needs a
second stage because terrain and body parameters cannot be observed on hardware,
whereas an external hand load is recoverable from the rigid-body model at run
time, so its latent is built from a measured quantity rather than a regressed
surrogate. Pose curricula and adversarial upper bodies perturb the stance with
motion the robot itself generates and can therefore predict from its own action
history; an external load is neither commanded nor anticipated. The same load produces a different base moment depending on the arm
configuration it acts through, which motivates encoding the two jointly rather
than as independent channels. Sec.~\ref{sec:q2} measures what that encoding
contributes over supplying the same force unencoded.


\section{Methodology}
\label{sec:methodology}
FAME conditions a base standing policy on a learned latent representation of
upper-body context. As illustrated in Fig.~\ref{fig:methodology}, the system
has three components: an upper-body context encoder, which maps joint states
and hand interaction forces to a latent vector describing the disturbance
bimanual manipulation induces; a base standing policy, which that latent
conditions; and a model-based estimator, which supplies the hand forces at
deployment. Training exposes the policy to diverse disturbances by sampling
external forces at the hands (Sec.~\ref{subsec:up_context_encoder}) and
randomizing upper-body targets through a pose curriculum that expands as
standing quality improves (Sec.~\ref{subsec:upper_body_curriculum}).
Training supplies the hand force exactly, whereas a deployed robot has no
such signal; we close that gap with a rigid-body estimator
(Sec.~\ref{subsec:force_estimation}) that requires no wrist sensing and no
supervision, and every result we report runs with the estimate in the loop.

\subsection{Upper-Body Pose Curriculum}
\label{subsec:upper_body_curriculum}

To train the standing controller under continuously varying upper-body
configurations, we employ the pose curriculum of HOMIE
\cite{BenQ-RSS-25} unchanged, which
progressively widens the range of randomized target poses as standing quality
improves. We maintain a scalar \emph{upper-body action ratio}
$\rho_a \in [0,1]$, initialized to $0$ and increased by $\Delta\rho$ whenever
the height-tracking reward exceeds a threshold. At each resampling event we
draw an auxiliary ratio $\rho_a'\in[0,1]$ that concentrates near zero early in
training and broadens as $\rho_a$ grows,
\begin{equation*}
\rho_a' = -\frac{1}{\kappa(1-\rho_a)}
\ln\!\Big(1-u\big(1-\exp(-\kappa(1-\rho_a))\big)\Big),
\label{eq:rhoa_simple}
\end{equation*}
where $u\sim\mathcal{U}(0,1)$ and $\kappa=20$ controls sharpness.
Let
$q_{\mathrm{ub}}\in\mathbb{R}^{15}$ denote the torso and arm joint positions. For each joint $j$ we sample
$a_j \sim \mathcal{U}(0,\rho_a')$ and a target angle in a curriculum-scaled
interval around the default posture $q^0_{\mathrm{ub}}$,
\begin{equation*}
q^{\star}_{\mathrm{ub},j}\sim
\mathcal{U}\!\Big(
q^0_{\mathrm{ub},j}-a_j\big(q^0_{\mathrm{ub},j}-q^{\min}_{\mathrm{ub},j}\big),\;
q^0_{\mathrm{ub},j}+a_j\big(q^{\max}_{\mathrm{ub},j}-q^0_{\mathrm{ub},j}\big)
\Big),
\label{eq:ub_target_simple}
\end{equation*}
resampled every $T_{\mathrm{ub}}=1$\,s. The evaluation poses of
Sec.~\ref{sec:sim_protocol} are drawn from this same distribution, so the
reported results measure behaviour inside the trained envelope rather than
extrapolation beyond it.

\subsection{Upper-Body Context Encoder}
\label{subsec:up_context_encoder}

The disturbance a manipulation task imposes on stance is not determined by the
hand force alone: the same force produces a different base moment depending on
the arm configuration through which it acts. We therefore encode the two
jointly rather than presenting either in isolation. The encoder $\mu_\theta$,
a multi-layer perceptron, 
maps the torso and arm
joint positions $q_{\mathrm{ub}}\in\mathbb{R}^{15}$ together with the two hand
forces $\mathbf{F}_L,\mathbf{F}_R\in\mathbb{R}^{3}$ to a latent context vector
\begin{equation*}
\mathbf{z}_{t}
= \mu_{\theta}\!\left(\left[q_{\mathrm{ub}},\,\mathbf{F}_L,\,\mathbf{F}_R\right]\right)
\in \mathbb{R}^{8},
\qquad
\left[q_{\mathrm{ub}},\,\mathbf{F}_L,\,\mathbf{F}_R\right]\in\mathbb{R}^{21}.
\label{eq:encoder_mapping}
\end{equation*}
It is trained end-to-end with the policy, with no force labels or auxiliary
regression loss, and the policy receives the $H_z=3$ most recent latents
appended to its proprioceptive history. During training the applied force is
drawn isotropically, $\mathbf{F} = r\,\bm{\xi}/\|\bm{\xi}\|_2$ with
$\bm{\xi}\sim\mathcal{N}(\mathbf{0},\mathbf{I}_3)$ and magnitude
$r\sim\mathcal{U}(r_{\min},r_{\max})$, giving unbiased coverage of the 3D
force space.

\subsection{Estimating Hand Force at Deployment}
\label{subsec:force_estimation}

At deployment, the hand force is not given, and recovering it from the
full-body dynamics is obstructed by the ground reaction, which is unknown and
discontinuous at contact events. We remove it structurally. Let $\mathcal{A}$
index the seven degrees of freedom of one arm; restricted to those rows, the
equation of motion reads
\begin{equation*}
\bigl[\mathbf{M}(q)\ddot{q} + \mathbf{h}(q,\dot{q})\bigr]_{\mathcal{A}}
= \bm{\tau}_{\mathcal{A}}
+ \bigl[\mathbf{J}_h^{\!\top} \mathbf{F}\bigr]_{\mathcal{A}}
+ \sum_{c} \bigl[\mathbf{J}_c^{\!\top} \mathbf{f}_c\bigr]_{\mathcal{A}},
\end{equation*}
where $\mathbf{J}_h$ is the Jacobian of the loaded point on the hand and
$\mathbf{J}_c$ that of a foot contact carrying the unknown reaction
$\mathbf{f}_c$. Moving an arm joint does not displace either foot, so the
columns of every $\mathbf{J}_c$ at the rows $\mathcal{A}$ are identically zero
and the contact sum vanishes: the ground reaction leaves the problem without
being measured or modelled.
The same argument decouples the two arms.
Evaluating the left-hand side rigid-body inverse dynamics with the recursive Newton--Euler algorithm and accounting for the actuator-side rotor inertia $\mathbf{B}$ and joint damping $\mathbf{D}$ gives
\begin{equation*}
\bigl[\mathbf{J}_{h}^{\!\top} \mathbf{F}\bigr]_{\mathcal{A}}
= \underbrace{\mathrm{RNEA}(q,\dot{q},\ddot{q})_{\mathcal{A}}
+ \mathbf{B}_{\mathcal{A}}\ddot{q}_{\mathcal{A}}
+ \mathbf{D}_{\mathcal{A}}\dot{q}_{\mathcal{A}}
- \bm{\tau}_{\mathcal{A}}}_{\textstyle \mathbf{r}_{\mathcal{A}}}.
\end{equation*}
For the 7-DoF arm, this gives seven scalar equations for the three components of the hand force.
We therefore estimate $\hat{\mathbf{F}}$ by solving the resulting overdetermined system in the least-squares sense and pass the result to the encoder.
Both $\mathbf{J}_h$ and the RNEA term are evaluated online with
Pinocchio \cite{carpentier2019pinocchio}. 


\subsection{Base policy}

We formulate the base-policy training problem as a Partially Observable
Markov Decision Process (POMDP)
\begin{equation*}
\mathcal{M} = \bigl(\,\mathcal{S},\;\mathcal{A},\;T,\;\mathcal{O},\;R,\;\gamma\bigr)
\end{equation*}
and utilize Proximal Policy Optimization (PPO)
\cite{schulman2017proximalpolicyoptimizationalgorithms} to learn a
parameterized policy $\pi_\theta(a_t \mid o_t)$ that maximizes the expected
return
\begin{equation*}
\label{exp_return}
\displaystyle J(\theta)=\mathbb{E}_{\pi_\theta}\bigl[\sum_t\,r_t\bigr] .
\end{equation*}
The policy observation concatenates a three-step proprioceptive history with
the latent context of Sec.~\ref{subsec:up_context_encoder},
\begin{equation*}
o_t = \bigl[\,\mathbf{p}_{t},\mathbf{p}_{t-1},\mathbf{p}_{t-2},\;
\mathbf{z}_{t},\mathbf{z}_{t-1},\mathbf{z}_{t-2}\,\bigr] ,
\end{equation*}
where $\mathbf{p}_t$ collects base angular velocity, projected gravity,
command, joint positions, joint velocities and the previous action; the state
space is detailed in Sec.~\ref{appendix:method_details/state_space}. We implement PPO from the
RSL-RL library \cite{schwarke2025rslrl}. Table~\ref{tab:standing_reward} summarizes the
rewards used to train the policy for the Unitree H1-2.

\begin{table}[!htb]
\resizebox{\linewidth}{!}{%
\renewcommand{\arraystretch}{1.1}
\centering
\begin{threeparttable}
\begin{tabular}{c c c c}
\hline
\textbf{Term} & \textbf{Expression} & \textbf{Weight} & \textbf{Remarks} \\
\hline
\rowcolor{lightgray}
\multicolumn{4}{c}{\textbf{Base Stability \& Uprightness}} \\
\hline

Base height tracking 
& $\exp(-4 |h_\text{base} - h_\text{cmd}|)$ 
& $3.0$ & $h_\text{cmd}=1.0$ \\

Vertical velocity 
& $v_z^2$
& $-5.0$ & Penalize bouncing \\

Angular velocity (xy) 
& $\|\boldsymbol{\omega}_{xy}\|_2^2$
& $-0.1$ & Suppress roll/pitch motion \\

Orientation 
& $\|\boldsymbol{g}_{xy}\|_2^2$
& $-3.0$ & Projected gravity error \\

Stand still (zero command) 
& $\mathds{1}(\text{motion at zero cmd})$
& $-0.8$ & Encourage static stability \\

\rowcolor{lightgray}
\hline
\multicolumn{4}{c}{\textbf{Posture \& Joint Deviation}} \\
\hline

Hip deviation 
& $\|\boldsymbol{q}_\text{hip} - \boldsymbol{q}_\text{hip}^0\|_2^2$
& $-0.2$ \\

Ankle deviation 
& $\|\boldsymbol{q}_\text{ankle} - \boldsymbol{q}_\text{ankle}^0\|_2^2$
& $-0.5$ \\

Knee deviation 
& $| (d_\text{knee}-0.5) \cdot (h_\text{base}-h_\text{cmd}) |$
& $-1.5$ & Height-coupled shaping \\

Joint tracking error 
& $\|\boldsymbol{q}_\text{target} - \boldsymbol{q}\|_2^2$
& $-0.1$ \\

DoF acceleration 
& $\|\ddot{\boldsymbol{q}}\|_2^2$
& $-2.5\times10^{-5}$ \\

\rowcolor{lightgray}
\hline
\multicolumn{4}{c}{\textbf{Feet Contact \& Grounding}} \\
\hline

Feet lateral distance 
& $\text{clamp}(d_\text{feet})$
& $0.5$ & Maintain stance width \\

Knee lateral distance 
& $\text{clamp}(d_\text{knee})$
& $1.0$ \\

Feet parallel 
& $\mathrm{Var}(d_\text{feet})$
& $-2.5$ & Symmetry constraint \\

Feet ground parallel 
& $\mathrm{Var}(h_\text{feet})$
& $-2.0$ \\

Feet slip 
& $\|\boldsymbol{v}_{\text{feet},xy}\|_2$
& $-1.0$ \\

Feet stumble 
& $\mathds{1}(F^{xy} > 3F^z)$
& $-1.5$ \\

Feet contact forces 
& $\max(\|F\|-900,0)$
& $-2.5\times10^{-4}$ \\

Contact momentum 
& $v_z^- \cdot (F_z-50)$
& $2.5\times10^{-4}$ & Encourage soft landing \\

No-fly (single contact) 
& $\mathds{1}(\text{single foot contact})$
& $0.75$ \\

\rowcolor{lightgray}
\hline
\multicolumn{4}{c}{\textbf{Control Smoothness \& Energy}} \\
\hline

Action rate 
& $\|\boldsymbol{a}_t - \boldsymbol{a}_{t-1}\|_2^2$
& $-0.02$ \\

Second-order smoothness 
& $\|\boldsymbol{a}_t - 2\boldsymbol{a}_{t-1} + \boldsymbol{a}_{t-2}\|_2^2$
& $-0.1$ \\

Joint power 
& $\sum |\dot{q}_i \tau_i|$
& $-2\times10^{-5}$ \\

Torque penalty 
& $\|\boldsymbol{\tau}\|_2^2$
& $-2.5\times10^{-6}$ \\

Action vanish 
& $\text{action saturation error}$
& $-1.0$ \\

\rowcolor{lightgray}
\hline
\multicolumn{4}{c}{\textbf{Safety \& Limits}} \\
\hline

DoF position limits 
& $\text{limit violation}$
& $-2.0$ & Soft limit = 0.975 \\

DoF velocity 
& $\|\dot{\boldsymbol{q}}\|_2^2$
& $-5\times10^{-3}$ \\

DoF velocity limits 
& $\text{soft vel violation}$
& $-2\times10^{-3}$ \\

Torque limits 
& $\text{soft torque violation}$
& $-0.1$ & Soft limit = 0.95 \\

\hline
\end{tabular}
\end{threeparttable}
}
\vspace{2mm}
\caption{Standing reward components and weights, adjusted from \cite{BenQ-RSS-25} for Unitree H1-2.}
\label{tab:standing_reward}
\end{table}



The policy outputs a 12D action $\bm{a}_t \in \mathbb{R}^{12}$, corresponding to target joint position offsets for the lower-body joints (6 per leg).
We compute target joint positions as
\begin{equation*}
\bm{q}^{\mathrm{tar}}_{t} = \bm{q}^{0}_{\mathrm{lb}} + s_a \, \bm{a}_t
\end{equation*}
where $\bm{q}^{0}_{\mathrm{lb}}$ is the default lower-body posture and $s_a$ is an action scaling factor.
Targets are tracked by a joint-space PD controller running at 50\,Hz:
\begin{equation*}
\bm{\tau}_t = \mathbf{K}_p\!\left(\bm{q}^{\mathrm{tar}}_{t}-\bm{q}_{t}\right)
            + \mathbf{K}_d\!\left(\dot{\bm{q}}^{\mathrm{tar}}_{t}-\dot{\bm{q}}_{t}\right)
\end{equation*}
with PD gains $\mathbf{K}_p,\mathbf{K}_d$ and  $\dot{\bm{q}}^{\mathrm{tar}}_{t}=\bm{0}$.

\begin{table*}[!htb]
\vspace{5pt}
\centering
\small
\setlength{\tabcolsep}{6pt}
\renewcommand{\arraystretch}{1.1}
\begin{tabular}{@{}l cccc cc c@{}}
\hline
 & \multicolumn{4}{c}{\textbf{Information available}}
 & \multicolumn{2}{c}{\textbf{Task success} (\%)}
 & \textbf{Deviation} \\
\cmidrule(lr){2-5}\cmidrule(lr){6-7}\cmidrule(lr){8-8}
\textbf{Variant}
 & \makecell{Pose\\curr.} & \makecell{Force to\\policy}
 & \makecell{Arm cfg.\\in encoder} & \makecell{Latent\\encoding}
 & $T{=}150$\,mm & $T{=}250$\,mm
 & $\Delta$EE (mm) \\
\hline
\textbf{FAME-pose} (HOMIE-style~\cite{BenQ-RSS-25}) & \checkmark & --- & \checkmark & \checkmark & 4.3 & 5.6 & 242.4 \\
\textbf{+Curr+F}$_\mathrm{raw}$ & \checkmark & \checkmark & --- & --- & 16.6 & 21.7 & 220.3 \\
ALMI~\cite{shi2025almi} & \multicolumn{4}{c}{\emph{adversarial motion imitation; no force input}} & 24.7 & 39.3 & 202.3 \\
\hline
\textbf{FAME} & \checkmark & \checkmark & \checkmark & \checkmark & \textbf{38.9} & \textbf{48.5} & \textbf{179.6} \\
\hline
\end{tabular}
\vspace{1mm}
\caption{Standing under swept hand-force disturbances over $100$ upper-body
configurations sampled from the training pose distribution (protocol in Sec.~IV-A).
A trial succeeds if the robot stays upright \emph{and} peak end-effector deviation
remains below $T$; poses a policy cannot hold count as failures rather than
exclusions. $\Delta$EE is the force-induced hand displacement at $T{=}250$\,mm,
differenced against the same trial run without force. ALMI is a separately trained locomotion policy, whose lower body is optimized against upper-body motion its own arms generate. All controlled variants receive the full 27-joint proprioceptive history. }
\label{tab:success_rates_cfg}
\end{table*}

\subsection{Training Variants}
\label{sec:variants}
We train three variants that differ only in what upper-body context reaches the
policy. \textbf{FAME-pose} encodes the arm configuration alone,
$z_t = \mu(q_{\mathrm{ub}})$; \textbf{F}$_\mathrm{raw}$ concatenates the two hand
forces into the observation without encoding; \textbf{FAME} encodes both jointly,
$z_t = \mu([q_{\mathrm{ub}}, F^L, F^R])$. \textbf{FAME-pose} against \textbf{FAME}
isolates the force channel, since both compress upper-body context to a latent of
the same width and differ only in whether force is part of it;
\textbf{F}$_\mathrm{raw}$ against \textbf{FAME} isolates the representation, since
both receive the force and differ only in whether it is concatenated directly or
compressed jointly with the arm configuration. All three share the pose curriculum,
reward function, force sampling, domain randomization, network widths, optimizer
settings, and step budget, and the concatenated variant is given the same history
depth $H_z=3$ as the latent, so that each comparison isolates the representation
rather than the temporal context available to the policy.

\section{Experiments}
\label{sec:experiments}

Our evaluation addresses three questions. \textbf{Q1:} does conditioning on
force configuration context expand the range of hand loads under which the
robot can keep standing and hold its hands where the task put them?
\textbf{Q2:} is the latent encoding responsible for that gain, or is the
availability of force information sufficient on its own? \textbf{Q3:} does the
result transfer beyond the environment in which it was trained to a different
simulator and task set, and to hardware, where the force the policy
conditions on is no longer exact?

\subsection{Simulation Protocol}
\label{sec:sim_protocol}
We sample $100$ upper-body configurations from the training pose distribution
of Sec.~\ref{subsec:upper_body_curriculum} at full curriculum. Each configuration is held fixed for the trial while
the lower body stands, and receives $25$ force draws: a per-hand horizontal
magnitude $r \sim \mathcal{U}(0,20)$\,N shared by both hands, an independent
uniform horizontal direction at each hand, and an independent vertical
component $F_z \sim \mathcal{U}(-30,30)$\,N. Spreading trials along the force
axis rather than repeating one magnitude is deliberate: the
between-configuration term dominates the standard error and does not shrink
with trials per configuration. This gives $2500$ loaded trials per policy over
a $10$\,s, with the first $2$\,s discarded as settling. The hand force
supplied to the encoder is the estimate of Sec.~\ref{subsec:force_estimation}.

\paragraph{Task-level error} Standing is a means, not the task. A controller
that stays upright but lets the hands drift has still failed the manipulation
it was standing for. We therefore run each trial twice, once with the force,
once without, from the same initial state, and difference the two hand trajectories pointwise.
\begin{equation*}
\Delta\mathrm{EE}(t) = p_{\mathrm{hand}}^{\mathrm{force}}(t)
- p_{\mathrm{hand}}^{\mathrm{no\text{-}force}}(t).
\end{equation*}
Everything that is not the disturbance cancels: the commanded height, the
baseline drift, the settling transient. What remains is what the load did to
the hand. The pairing also makes policies with different command interfaces
directly comparable, since any term common to both runs vanishes.

\paragraph{Success criterion} A trial succeeds if the robot remains upright
\emph{and} the peak $\Delta\mathrm{EE}$ over the horizon stays below a task
tolerance $T$; we report $T = 150$\,mm and $T = 250$\,mm, spanning the
placement accuracy a bimanual carry or door-opening task would demand. A fall
is a task failure, not a missing measurement, so falls are censored at $T$
rather than excluded. A policy that fails often is penalised rather than
scored only on the trials it survived. Because the task is standing, a policy
that recovers by walking out from under the load has also failed it: trials
displacing either foot by more than $0.15$\,m are counted as failures.
Q1 and Q2 share this protocol and are reported together in
Table~\ref{tab:success_rates_cfg}.

\begin{figure*}[!ht]
    \vspace{4pt}
    \centering
    \includegraphics[width=0.93\textwidth]{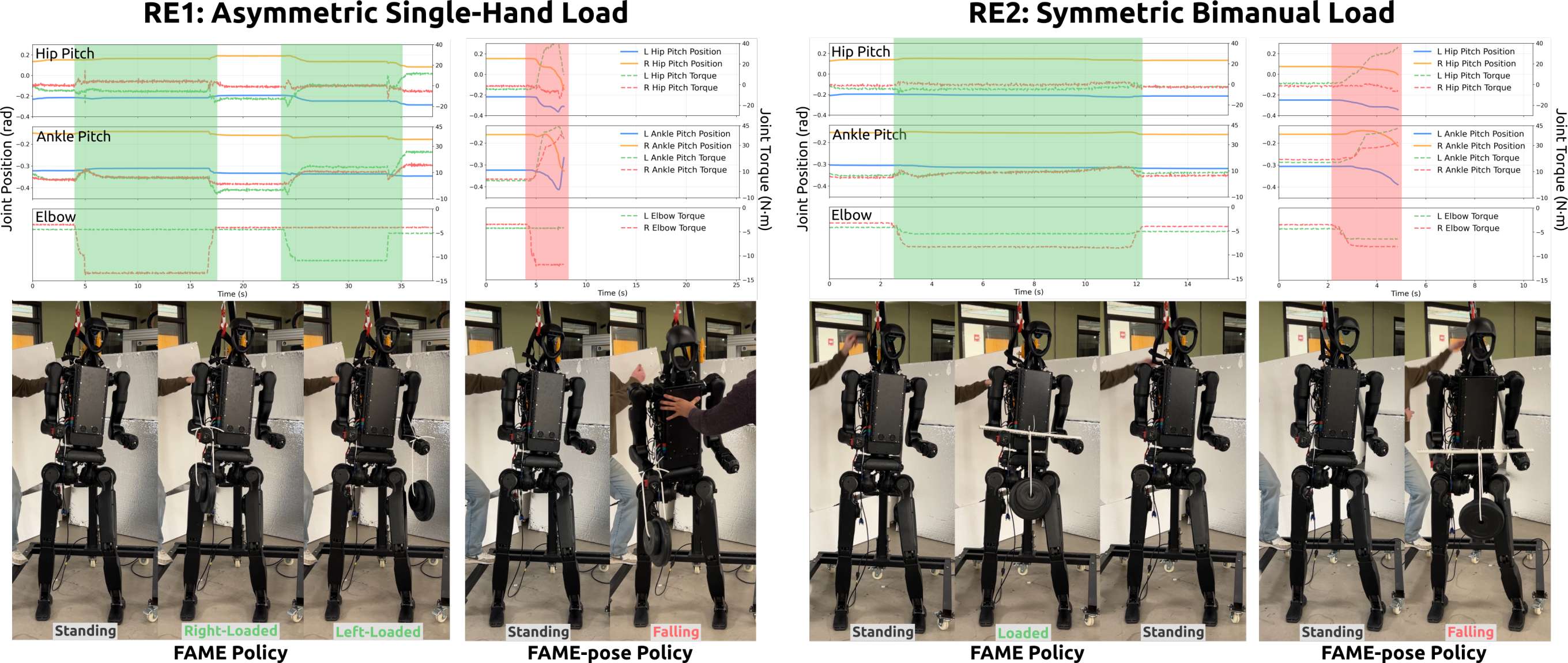}
    \vspace{-2pt}
    \caption{\textbf{Real-robot qualitative results.} Snapshot sequence from our real-robot evaluation on the Unitree H1-2 under representative load-interaction disturbances (RE1--RE2). For each experiment, we report the joint trajectories and torques of the hip pitch, ankle pitch, and elbow joints. With FAME, the robot remains stable under external loads and the joint positions stay close to their nominal standing configuration (marked in \textcolor{green}{green}). With FAME-pose, the joint positions drift away from the stable configuration, ultimately causing the robot to lose balance and fall (marked in \textcolor{red}{red}).}
    \label{fig:real_robot_snapshots}
    \vspace{-2mm}
\end{figure*}

\subsubsection{Q1: Envelope Expansion}
\label{sec:q1}
\textbf{FAME-pose} and \textbf{FAME} differ only in whether the interaction
force enters the encoded context, so the contrast isolates the force channel.

\textbf{Results and Discussion.} Force conditioning raises task success from
$4.3\%$ to $38.9\%$ at $T{=}150$\,mm and from $5.6\%$ to $48.5\%$ at
$T{=}250$\,mm, close to a ninefold gain at both tolerances. Deviation moves
with it: peak displacement falls from $242.4$ to $179.6$\,mm, so the trials
FAME does not convert into successes it still loses by a smaller margin. The pose curriculum alone therefore leaves most of the admissible envelope unreached, which shows what the policy needs is not further exposure to arm configurations but knowledge of the load acting through them.

\subsubsection{Q2: Encoding versus Raw Force Input}
\label{sec:q2}
Sec.~\ref{sec:q1} establishes that a force-conditioned policy outperforms one
without force information, but not whether the latent encoding contributes
beyond the availability of that information.
\textbf{+Curr+F}$_\mathrm{raw}$ versus \textbf{FAME} resolves this: both
receive the interaction force and differ only in whether it is concatenated
into the observation or compressed jointly with the arm configuration into a
latent of fixed width. Both are given the same history depth, so the comparison
is of representation rather than of temporal context.

\textbf{Results and Discussion.} Availability of force is not sufficient.
+Curr+F$_\mathrm{raw}$ reaches $16.6\%$ and $21.7\%$, well ahead of
FAME-pose's $4.3\%$ and $5.6\%$. The force signal plainly carries usable
information, but under half of FAME's $38.9\%$ and $48.5\%$. Its deviation,
$220.3$\,mm, sits nearer FAME-pose's $242.4$ than FAME's $179.6$. These results indicate that much of the gain arises from jointly representing force and arm configuration rather than simply providing the force signal. The result also reflects on the
co-trained estimate design of \cite{10.1007/978-981-95-2095-4_5}, the gain
we measure comes from the representation of the force, not merely from having
obtained one.

\textbf{Comparison with adversarially trained locomotion
ALMI \cite{shi2025almi}}: It acquires robustness without representing the
disturbance. Its lower body is trained against upper-body motion that its own
arms generate, and its observation carries no force channel, so the comparison
tests whether robustness to endogenous disturbance transfers to sustained
exogenous hand loading. It transfers partially. ALMI reaches $24.7\%$ task
success at $T{=}150$\,mm and $39.3\%$ at $T{=}250$\,mm, ahead of every variant
that lacks the joint encoding but behind FAME's $38.9\%$ and $48.5\%$, and its
deviation of $202.3$\,mm exceeds FAME's $179.6$\,mm. The gap narrows as the
tolerance widens: ALMI gains $14.6$ points between the two tolerances where
FAME gains $9.6$, indicating that its failures are concentrated as near-misses
in that band. It keeps the robot upright without keeping the hands still. This behavior is suitable for locomotion recovery but violates the fixed-stance requirement of the manipulation setting.

\begin{figure*}[!t]
  \centering
  \includegraphics[width=0.85\textwidth]{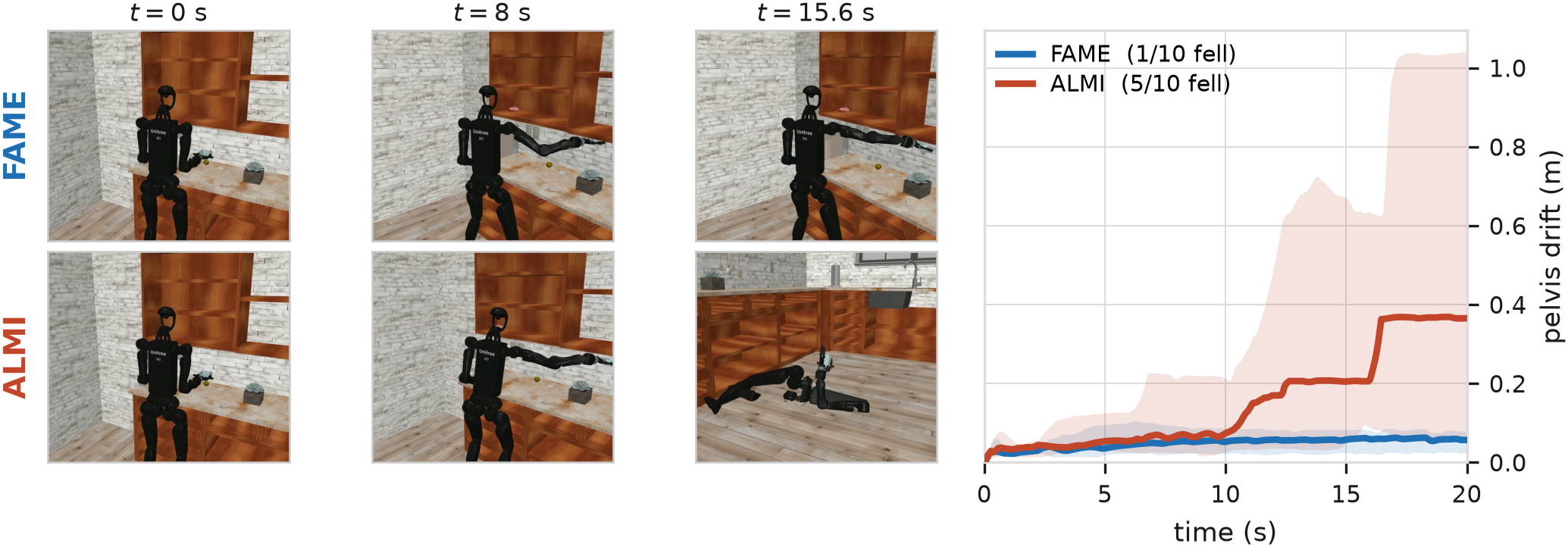}
\caption{Task-level comparison in the GOLEM kitchen twin. Both policies perform
the same single-arm pick of a loaded object, with the upper body driven by a
shared inverse-kinematics stack so that the lower body is the only variable.
Left: one seed; FAME holds its stance, ALMI steps out from under
the load and falls. Right: pelvis drift over $10$ seeds.}

  \label{fig:golem}
\end{figure*}

\subsubsection{Q3: Transfer Beyond the Training Environment}
\label{sec:q3}
\textbf{Sim-to-sim: task-level manipulation}
Forces produced by a task are neither isotropic nor independent of the robot's
own motion, unlike the sampled disturbances of Sec.~\ref{sec:sim_protocol}. We
therefore repeat the comparison on the GOLEM
platform~\cite{conway2026golemmodularhumanoidautonomy}, which exposes FAME and
ALMI behind a common velocity-command interface in a MuJoCo kitchen twin, with
the upper body driven by a shared inverse-kinematics stack so that the
comparison isolates lower-body behaviour. The task, a single-arm pick of a
$3$\,kg object from the counter, is shown in Fig.~\ref{fig:golem}; we run $10$
seeds varying the object and its spawn pose, both policies on every seed. Additional kitchen interaction in a different environment, the opening of the refrigerator door (Fig~\ref{fig:overview}, bottom row), is qualitatively demonstrated.

\textbf{Results and Discussion.}
Both policies hold the stance while the arm reaches and closes on the object.
FAME then falls on $1$ of $10$ seeds against ALMI's $5$, with median peak pelvis
displacement of $0.065$\,m against $0.641$\,m and median peak tilt of
$7.1^\circ$ against $30.5^\circ$. ALMI yields to the load and repositions its
feet, correct for a locomotion controller and wrong when the feet are where the
task placed them. The seed FAME fails is one ALMI fails too.

\textbf{Sim-to-real: hardware deployment.}
We deploy on the Unitree H1-2, which has
$27$ degrees of freedom: two $7$-DoF arms, two $6$-DoF legs, and a $1$-DoF
waist. Base orientation and angular velocity come from the onboard IMU, joint
angles and velocities from the joint encoders; no wrist force/torque sensor is
used. The base policy runs in real time at $50$\,Hz. We evaluate two scenarios: \textbf{RE1: Asymmetric Single-Arm Load} applies a load through one arm, producing an unbalanced moment about the stance. \textbf{RE2: Symmetric
 bimanual} load distributes the weight across both arms. 

\textbf{Results and Discussion.} As shown in Fig.~\ref{fig:real_robot_snapshots}, FAME held its stance in 3/3 trials of both RE1 and RE2, while FAME-pose lost balance and fell in 3/3 of each. The difference is clearest in the hip-pitch and ankle-pitch trajectories: FAME regulates joint torque to hold these joints near the nominal
standing posture, whereas FAME-pose does not account for the external
load and the joint positions drift until the balanced pose is lost.


\section{Conclusion}
\label{sec:conclusion}
We presented FAME, a force-adaptive RL framework that expands the manipulation
envelope of a full-scale humanoid by conditioning standing control on a latent
context encoding upper-body configuration and bimanual interaction force
jointly, with the force reconstructed online from joint torques rather than
measured. Over $100$ arm configurations sampled from the training pose
distribution and scored by load-induced displacement,
FAME reaches $38.9\%$ task success at a $150$\,mm tolerance, against $16.6\%$
for a policy given the same estimated force without the encoding, $4.3\%$ for a
pose-conditioned curriculum policy, and $24.7\%$ for an adversarially trained
locomotion policy. The first contrast holds the force input fixed and varies only its representation, attributing the gain to the joint encoding. The same estimator runs unchanged on kitchen manipulation tasks and on the
Unitree H1-2, where the advantage persists despite a noisier estimate.
Extending the joint encoding to load-carrying locomotion, where the feet may
move but the hands must still arrive, is the natural next step.


\section*{ACKNOWLEDGMENT}
This work has been supported by ARPA-E contract DE-AR0001966. We are grateful for this support. We leverage generative AI tools (Opus-5) for structural and grammatical proofreading in all sections and for assistance in generating code for plots in this manuscript.  N. Correll is CEO of Realtime Manufacturing, Inc., which is working on humanoids for manufacturing applications. 

{
    \small

    \bibliography{main}
    
}
\appendix


\subsection{State Space Design}
\label{appendix:method_details/state_space}


\textbf{Actor (policy) observations.}
The actor observation at time $t$ is a stacked history of length $H_a=3$:
$
\bm{o}^{\pi}_t = [\bm{o}_t, \bm{o}_{t-1}, \bm{o}_{t-2}],
$
where each one-step vector $\bm{o}_t \in \mathbb{R}^{d_o}$ is defined in Table~\ref{tab:actor_state}.
\begin{table}[!htb]
\centering
\footnotesize
\setlength{\tabcolsep}{3pt}
\renewcommand{\arraystretch}{0.95}
\begin{tabular}{@{}p{0.60\columnwidth} c c@{}}
\hline
\textbf{Actor observation term (one-step)} & \textbf{Symbol} & \textbf{Dim} \\
\hline
Command (scaled) & $\bm{c}_t$ & 3 \\
Commanded base height & $h^{\text{cmd}}_t$ & 1 \\
IMU angular velocity (body frame) & $\bm{\omega}_t$ & 3 \\
Projected gravity (body frame) & $\bm{g}_t$ & 3 \\
Joint position error (scaled) & $\tilde{\bm{q}}_t$ & $N_{\text{dof}}$ \\
Joint velocity (scaled) & $\tilde{\dot{\bm{q}}}_t$ & $N_{\text{dof}}$ \\
Previous action & $\bm{a}_{t-1}$ & $N_{\text{act}}$ \\
\hline
\multicolumn{3}{@{}l@{}}{$d_o = 3{+}1{+}3{+}3{+}2N_{\text{dof}}{+}N_{\text{act}} = 76$ \;($N_{\text{dof}}{=}27$, $N_{\text{act}}{=}12$)} \\
\multicolumn{3}{@{}l@{}}{$\dim(\bm{o}^{\pi}_t) = H_a d_o + H_z\cdot 8 = 228 + 24 = 252$ \;($H_a{=}H_z{=}3$)} \\
\hline
\end{tabular}
\caption{Actor (policy) observation space.}
\label{tab:actor_state}
\end{table}

\textbf{Critic (value) observations.}
The critic uses a privileged one-step vector $\bm{o}^V_t \in \mathbb{R}^{d_v}$ that augments $\bm{o}_t$ with the base linear velocity (available in simulation).
We use a history length $H_c=1$, i.e., $\bm{o}^V_t$ only, as summarized in Table~\ref{tab:critic_state}.
\begin{table}[!htb]
\centering
\renewcommand{\arraystretch}{1.05}
\begin{tabular}{l c c}
\hline
\textbf{Critic observation term} & \textbf{Symbol} & \textbf{Dim} \\ \hline

All actor one-step terms & $\bm{o}_t$ & 76 \\
Latent context (from encoder) & $\bm{z}_t$ & 8 \\
Base linear velocity (privileged, scaled) & $\bm{v}^{\text{base}}_t$ & 3 \\ \hline

\multicolumn{3}{l}{\textbf{One-step critic obs: } $d_v = d_o + 8 + 3 = 87$} \\
\multicolumn{3}{l}{\textbf{History stacking: } $H_c = 1 \Rightarrow \dim(\bm{o}^{V}_t)= 87$} \\ \hline
\end{tabular}
\vspace{2mm}
\caption{Critic observation design. The critic is additionally provided latent context $\bm{z}_t$ and privileged base linear velocity in simulation during training.}
\label{tab:critic_state}
\end{table}

\subsection{Domain Randomization}
\label{appendix:method_details/domain_rand}

Table \ref{tab:domain_rand_standing} summarizes the domain randomization strategies that we used.
\begin{table}[!htb]
\centering
\footnotesize
\renewcommand{\arraystretch}{0.95}
\setlength{\tabcolsep}{3pt}
\begin{tabular}{@{}>{\centering\arraybackslash}p{0.29\columnwidth} >{\RaggedRight\arraybackslash}p{0.63\columnwidth}@{}}
\hline
\textbf{Term} & \textbf{Value} \\
\hline
\rowcolor{lightgray}
\multicolumn{2}{@{}c@{}}{\textbf{External Disturbances}} \\
Push robot & $\text{interval}=4\,\mathrm{s}$, $v_\mathrm{xy}\sim\mathcal{U}(0,0.5)\,\mathrm{m/s}$ \\
Upper-body disturbance curriculum & $\rho_a\leftarrow 0.0$ initially; increased with standing quality, interval $=1\,\mathrm{s}$ \\
\rowcolor{lightgray}
\multicolumn{2}{@{}c@{}}{\textbf{Actuation \& Observation Noise}} \\
Joint injection noise & $q\leftarrow q+\epsilon$, $\epsilon\sim\mathcal{U}(-0.05,0.05)$ \\
Actuation offset & $a\leftarrow a+\delta$, $\delta\sim\mathcal{U}(-0.05,0.05)$ \\
Control delay & enabled \\
\rowcolor{lightgray}
\multicolumn{2}{@{}c@{}}{\textbf{Inertial \& Body Parameter Randomization}} \\
Payload mass & $m_\text{payload}\sim\mathcal{U}(-3,5)\,\mathrm{kg}$ \\
Hand payload mass & $m_\text{hand}=0$ (disabled) \\
Body displacement & $\Delta\boldsymbol{p}_\text{body}\sim\mathcal{U}(-0.1,0.1)\,\mathrm{m}$ \\
Link mass scaling & $m_\text{link}\leftarrow s\,m_\text{default}$, $s\sim\mathcal{U}(0.8,1.2)$ \\
\rowcolor{lightgray}
\multicolumn{2}{@{}c@{}}{\textbf{Contact Randomization}} \\
Friction coefficient & $\mu\sim\mathcal{U}(0.1,3.0)$ \\
Restitution & $e\sim\mathcal{U}(0.0,1.0)$ \\
\rowcolor{lightgray}
\multicolumn{2}{@{}c@{}}{\textbf{Controller Randomization}} \\
Proportional gain & $k_p\leftarrow s\,k_{p,\text{default}}$, $s\sim\mathcal{U}(0.9,1.1)$ \\
Derivative gain & $k_d\leftarrow s\,k_{d,\text{default}}$, $s\sim\mathcal{U}(0.9,1.1)$ \\
Initial joint position & $q_0\leftarrow s\,q_{0,\text{default}}$, $s\sim\mathcal{U}(0.8,1.2)$; then $q_0\leftarrow q_0+\Delta q$, $\Delta q\sim\mathcal{U}(-0.1,0.1)$ \\
\hline
\end{tabular}
\caption{Domain randomizations used for training the standing policy. $\mathcal{U}(\cdot)$ denotes a uniform distribution.}
\label{tab:domain_rand_standing}
\end{table}

\subsection{Upper-Body Context Encoder \texorpdfstring{$(\mu)$}{(mu)} and Base Policy \texorpdfstring{$(\pi)$}{(pi)} Architectures}

In our framework, for the upper-body context encoder $\mu$, we use an MLP that maps upper-body pose and bimanual interaction forces to a latent context $z_t$, which conditions the actor for force-adaptive standing.
For the actor-critic, we use HIM-style \cite{long2024hybrid, BenQ-RSS-25}, where the actor consists of an estimator network $\mathcal{E}$ and a follow-up policy network $\mathcal{N}$, and the critic is a separate value network $\mathcal{C}$.
\lstset{
  backgroundcolor=\color{gray!10},
  basicstyle=\ttfamily\small,
  breaklines=true,
  frame=single,
  showstringspaces=false,
  mathescape=true
}

\subsection{Hyperparameters}
Table \ref{tab:hyperparam_standing} presents the hyperparameters used for training.
\begin{table}[!b]
\centering
\footnotesize
\setlength{\tabcolsep}{3pt}
\renewcommand{\arraystretch}{0.95}
\begin{tabular}{@{}llll@{}}
\toprule
\textbf{Hyperparameter} & \textbf{Value} & \textbf{Hyperparameter} & \textbf{Value} \\
\midrule
Optimizer          & Adam              & Discount factor $\gamma$ & 0.99  \\
Learning rate      & $1\times10^{-3}$  & GAE $\lambda$            & 0.95  \\
Schedule           & adaptive          & Clip param $\epsilon$    & 0.2   \\
Learning epochs    & 5                 & Value loss coef          & 1.0   \\
Mini-batches       & 4                 & Clipped value loss       & True  \\
Max grad norm      & 1.0               & Entropy coef             & 0.01  \\
Init noise std     & 1.0               & Desired KL               & 0.01  \\
Actor / critic MLP & [512, 256, 256]   & Symmetry scale           & 1.0   \\
Activation         & ELU               & & \\
\bottomrule
\end{tabular}
\caption{PPO hyperparameters used for training.}
\label{tab:hyperparam_standing}
\end{table}


\end{document}